\documentclass[runningheads]{llncs}

\usepackage{accv}

\usepackage{accvabbrv}

\usepackage{graphicx}
\usepackage{booktabs}
\usepackage{multirow}
\usepackage{amsmath}
\usepackage{amssymb}
\usepackage{xcolor}

\usepackage[accsupp]{axessibility}  

\usepackage{hyperref}

\usepackage{orcidlink}

\begin{document}

\title{ImCorr: Sub-pixel Semantic Correspondence via Implicit Feature Decoding} 


\author{Yusung Choi\inst{1}}

\authorrunning{Y.~Choi}

\institute{Pukyong National University, Busan, Republic of Korea \\
\email{cyscyb@gmail.com}}

\maketitle

\begin{abstract}
The strong performance that modern semantic correspondence methods achieve at standard thresholds plateaus sharply at fine-grained thresholds. We argue that this plateau stems not from the representational capacity of backbone features, but from a grid-tied readout. Patch-based vision transformers tokenize images onto discrete grids, introducing two forms of quantization error. On the source side, the nearest patch feature is queried in place of the exact keypoint coordinate. On the target side, no grid feature exists that represents the precise location of the ground truth. We quantify this quantization ceiling across all 499,188 keypoints in SPair-71k, the standard benchmark for semantic correspondence: under the standard $448 \times 448$, patch-14 setting, 84.9\% of ground-truth keypoints have no grid feature representing their precise location at PCK@0.01. This is a structural limitation at the representation level, independent of the matching strategy. We address this with \textbf{ImCorr: Sub-pixel Semantic Correspondence via Implicit Feature Decoding}, which formulates correspondence estimation over a continuous feature field queryable at arbitrary continuous coordinates. A FiLM-conditioned decoder is trained to embed sub-pixel positional information into the feature field. On the source side, querying the field directly at the exact keypoint coordinate theoretically reduces the representation-level quantization error to zero; on the target side, decoding the field onto a grid arbitrarily denser than the backbone grid substantially reduces quantization error. On SPair-71k and AP-10K (intra-species, cross-species, and cross-family), ImCorr improves performance at fine-grained thresholds (PCK@0.01--0.05), achieving a 6.2 percentage point gain over the prior state of the art at PCK@0.01 on SPair-71k. These results experimentally demonstrate that representational continuity is an effective solution for precise semantic correspondence. Code is available at \url{https://github.com/YusungChoi/ImCorr}.

  \keywords{Semantic Correspondence \and Implicit Feature Decoding \and Sub-pixel Localization}
\end{abstract}

\section{Introduction}

There are no grid lines in nature. The tip of a bird's wing, the edge of a car's side mirror, the corner of a human eye — the semantic correspondences we seek between images exist in a continuous space that is indifferent to pixel grids. Yet the models that estimate them decompose images into discrete patches, define features only at grid points, and search for answers only on the grid. We argue that this gap is the fundamental bottleneck of precise correspondence.

Semantic correspondence (SC) is the problem of finding semantically matching locations across different images, and serves as a foundation for downstream tasks requiring precise spatial reasoning, including pose estimation, scene understanding, and robotic manipulation. Recent methods leveraging large pretrained vision models — most notably DINOv2~\cite{oquab2023dinov2} and Stable Diffusion~\cite{rombach2022high} — as backbones have achieved remarkable gains on standard benchmarks. These advances have been driven by richer semantic representations, more sophisticated cost aggregation, and stronger transformation invariance~\cite{tang2023emergent,zhang2023tale,zhang2024telling}. Yet all such methods share a common assumption: features are defined on a grid, and correspondence search is conducted on the same grid.

\begin{figure*}[t]
    \centering
    \includegraphics[width=\textwidth]{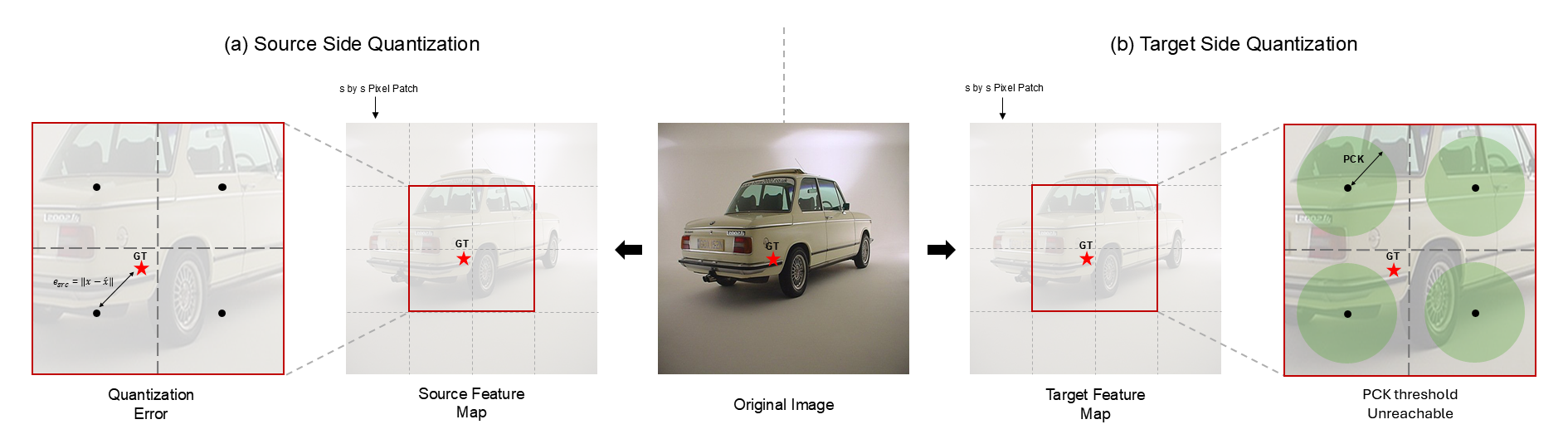}
    \caption{
        Illustration of grid quantization error in semantic correspondence.
        (a) \textbf{Source-side quantization}: the same patch feature is used in matching regardless of where within the patch a keypoint lies, so the keypoint's exact positional information is not reflected in the representation. The distance to the nearest patch center, $\|\mathbf{x} - \hat{\mathbf{x}}\|$, quantifies the extent of positional information disregarded.
        (b) \textbf{Target-side quantization}: when the ground-truth location falls between grid points, no grid feature represents its precise location, making correct localization structurally impossible within the PCK threshold.
    }
    \label{fig:quantization}
\end{figure*}

This assumption is benign at the standard threshold (PCK@0.1), where the permissible error is large relative to the grid spacing. At the fine-grained threshold (PCK@0.01), however, the situation changes fundamentally. The discreteness of the grid itself structurally prevents reaching the ground truth. We term this the \textbf{quantization ceiling} — an upper bound on achievable precision that arises not from inadequate backbone features, but from a grid-tied readout that cannot measure what the backbone knows. The error manifests in two places. On the source side, the feature of the nearest patch center, rather than the exact keypoint coordinate, is fed into matching, imprinting a positional error into the representation before any matching begins. On the target side, no grid feature exists that represents the precise location of the ground truth.

Table~\ref{tab:quantization_ceiling} quantifies this ceiling across all 499,188 keypoints in SPair-71k~\cite{min2019spair}. On the source side, the same patch feature is used in matching regardless of where within the patch a keypoint lies, meaning that the keypoint's exact positional information is not reflected in the representation. For patch sizes 14 and 16, the distance between a keypoint and its nearest patch center reaches mean values of 5.35px and 6.13px, and maximum values of 9.90px and 11.31px, respectively, irrespective of image resolution — quantifying the extent of positional information disregarded by the patch feature. On the target side, under the standard $448\times448$, patch-14 setting, 84.9\% of ground-truth keypoints have no grid feature representing their precise location at PCK@0.01; doubling the resolution still leaves 44.1\% in this state.

Increasing resolution or interpolating grid features offers a possible way to mitigate this problem. However, interpolation only blends existing grid values and does not generate new visual information between grid points.

\begin{table}[t]
\centering
\setlength{\tabcolsep}{8pt}
\begin{tabular}{ccccc}
\toprule
Image Size & Patch Size & PCK@0.1 & PCK@0.05 & PCK@0.01 \\
\midrule
\multirow{2}{*}{224$\times$224} & 14 & 2.70\% & 27.56\% & 96.23\% \\
                                 & 16 & 5.30\% & 36.72\% & 97.22\% \\
\midrule
\multirow{2}{*}{448$\times$448} & 14 & 0.26\% & 2.88\% & \textbf{84.91\%} \\
                                 & 16 & 0.28\% & 5.41\% & 88.59\% \\
\midrule
\multirow{2}{*}{896$\times$896} & 14 & <0.01\% & 0.22\% & \textbf{44.06\%} \\
                                 & 16 & 0.03\% & 0.34\% & 55.16\% \\
\bottomrule
\end{tabular}
\vspace{2mm}
\caption{Proportion of keypoints unreachable within the PCK threshold (relative to target bounding box) when only feature map grid points are considered as candidates (SPair-71k, 499,188 keypoints). This represents a structural ceiling independent of backbone quality.}
\label{tab:quantization_ceiling}
\vspace{-2mm}
\end{table}

To address this, we propose \textbf{ImCorr: Sub-pixel Semantic Correspondence via Implicit Feature Decoding}, which reformulates correspondence estimation over a continuous feature field $F(\mathbf{x})$, addressing this limitation at the representation level. A coordinate-conditioned FiLM decoder transforms the within-grid offset into channel-wise modulation~\cite{perez2018film}, directly learning sub-pixel positional information that interpolation cannot produce. On the source side, the continuous field is queried directly at the exact keypoint coordinate, theoretically eliminating the representation-level quantization error; on the target side, the feature field is decoded onto a grid denser than the backbone grid, substantially reducing quantization error.

Experiments on SPair-71k~\cite{min2019spair} and AP-10K~\cite{yu2021ap} show that ImCorr achieves consistently superior performance at fine-grained thresholds (PCK@0.05, PCK@0.01). These results experimentally demonstrate that grid quantization — long obscured by the loose thresholds of standard benchmarks — is in fact the bottleneck of precise correspondence, and that representational continuity is its solution. Our contributions are summarized as follows:

\begin{itemize}
\item We formalize and empirically demonstrate that the performance plateau at fine-grained thresholds originates from the structural limitations of grid-tied readout, rather than from insufficient backbone representations.

\item We propose ImCorr, which formulates semantic correspondence over a continuous feature field. Through a FiLM-conditioned decoder, the source side queries the field at exact keypoint coordinates to theoretically eliminate quantization error, while the target side decodes onto a grid denser than the backbone lattice to substantially reduce the proportion of unreachable candidates.

\item \item ImCorr achieves consistent performance gains at fine-grained thresholds (PCK @0.01, PCK@0.05) over existing methods on both SPair-71k and AP-10K (intra-species, cross-species, and cross-family).
\end{itemize}

\section{Related Work}

\paragraph{\textbf{Quantization in Semantic Correspondence.}} Semantic correspondence has long been a fundamental problem in computer vision. Early methods relied on handcrafted features such as HOG~\cite{dalal2005histograms} and SIFT~\cite{lowe2004distinctive}, before CNN-based representations became the dominant paradigm. More recently, large-scale pretrained vision models — most notably DINOv2~\cite{oquab2023dinov2} and Stable Diffusion~\cite{rombach2022high} — have been shown to provide powerful semantic representations, and methods built upon these backbones have achieved remarkable gains on standard benchmarks. SD-DINO fuses features from both models~\cite{zhang2023tale} to achieve strong zero-shot correspondence estimation, while GeoAware-SC enhances geometric awareness~\cite{zhang2024telling} to address challenging cases involving symmetric structures. More recently, MARCO combines a coarse-to-fine training objective with self-distillation~\cite{cuttano2026marco} to improve overall correspondence performance and generalization.

Despite these advances, all such methods share a common blind spot: they focus exclusively on improving the quality of features provided by the backbone, while leaving unquestioned the discreteness of the space in which those features are defined. SD-DINO~\cite{zhang2023tale}, GeoAware-SC~\cite{zhang2024telling}, and their contemporaries all extract features and search for correspondences on a patch-tokenized grid, and the fact that this design choice imposes a structural ceiling on precision at fine-grained thresholds has gone largely unnoticed.

A separate line of work attempts to produce continuous predictions from discrete grid features at the matching stage. Window soft-argmax~\cite{zhang2024telling}, as employed in GeoAware-SC, is one such approach, producing sub-grid predictions by computing a weighted average over neighboring grid features. However, this is a matcher-level operation designed to improve matching quality, and the discreteness of the underlying feature representation remains intact. In contrast, we address quantization not by modifying the matcher, but by operating on a continuous feature representation — one in which the features themselves are defined at arbitrary sub-pixel coordinates.

\paragraph{\textbf{Implicit Neural Representation.}} Implicit neural representations (INRs) are a paradigm for representing signals as continuous functions rather than discrete grids, and gained widespread attention in 3D vision through NeRF~\cite{mildenhall2021nerf}, which represents a 3D scene as a continuous radiance field queryable from arbitrary viewpoints, overcoming the resolution limitations of discrete voxel grids. This idea has since been extended to 2D vision: LIIF represents images as continuous functions~\cite{chen2021learning} from which pixel values can be queried at arbitrary resolutions, breaking free from the constraints of fixed grid resolution in super-resolution tasks.

Within semantic correspondence, NeMF was the first to introduce implicit neural representations to the field~\cite{hong2022neural}, representing the 4D matching cost between a source-target image pair as an implicit function. NeMF processes a coarse cost volume as guidance through a cost embedding network, and establishes correspondence through a subsequent fully-connected network. At inference, NeMF iteratively performs PatchMatch-based search and coordinate optimization to produce precise correspondences.

Our method differs fundamentally from NeMF in two respects. First, NeMF implicitly represents the matching cost~\cite{hong2022neural} while leaving the underlying feature representations grid-bound; our method instead defines a continuous per-image feature field, so that features themselves are defined at arbitrary sub-pixel coordinates. Second, NeMF requires iterative coordinate optimization at inference time, whereas our method produces correspondences in a single forward pass. Furthermore, whereas LIIF predicts pixel values for super-resolution~\cite{chen2021learning}, our continuous feature field directly targets the quantization bottleneck in semantic correspondence, learning to embed sub-pixel positional information into the feature representation itself.

\section{Method}
\subsection{Problem Formulation}
\label{sec:formulation}

Given a source image $\mathbf{I}_s \in \mathbb{R}^{H \times W \times 3}$ and a target image $\mathbf{I}_t \in \mathbb{R}^{H \times W \times 3}$, the goal of semantic correspondence is to estimate, for a keypoint coordinate $\mathbf{x} \in \mathbb{R}^2$ in the source image, the semantically corresponding coordinate $\mathbf{y} \in \mathbb{R}^2$ in the target image.

\paragraph{\textbf{Grid-based Matching Formulation.}}
Existing methods extract source and target feature maps $\mathbf{Z}_s, \mathbf{Z}_t \in \mathbb{R}^{H' \times W' \times C}$ using a backbone encoder. To achieve high-level semantic invariance, the backbone extracts features from deep layers, which reduces spatial resolution: a region of $s \times s$ pixels in the original image is compressed into a single $C$-dimensional feature vector, yielding a downsampled feature map of size $H' = H/s$, $W' = W/s$. Representative backbones such as DINOv2~\cite{oquab2023dinov2} and Stable Diffusion~\cite{rombach2022high} employ patch strides of $s \approx 14$ or $16$, meaning that a $14 \times 14$ or $16 \times 16$ pixel region is represented by a single vector.

The discrete grid $\mathcal{G} = \{(i \cdot s,\; j \cdot s) \mid i \in [0, H'),\; j \in [0, W')\}$ defines both the space in which features are defined and the set of candidate locations $\mathbf{p} \in \mathcal{G}$ for correspondence search. Correspondence estimation is then formulated as similarity maximization over this grid:
\begin{equation}
    \hat{\mathbf{y}} = \underset{\mathbf{p} \in \mathcal{G}}{\arg\max}\; \mathrm{sim}\!\left(\mathbf{Z}_s[\hat{\mathbf{x}}],\; \mathbf{Z}_t[\mathbf{p}]\right),
    \label{eq:grid_matching}
\end{equation}
where $\hat{\mathbf{x}} = \arg\min_{\mathbf{p} \in \mathcal{G}} \|\mathbf{x} - \mathbf{p}\|$ is the nearest grid point to the source keypoint $\mathbf{x}$, and $\mathrm{sim}(\cdot, \cdot)$ denotes cosine similarity.

This $s \times s$ compression imprints two levels of quantization error into the correspondence pipeline. On the source side, regardless of where keypoint $\mathbf{x}$ lies within a grid region, only the feature of the grid center $\hat{\mathbf{x}}$ is fed into matching. The resulting representation error is bounded by:
\begin{equation}
    e_\mathrm{src}(\mathbf{x}) = \|\mathbf{x} - \hat{\mathbf{x}}\| \leq \frac{s\sqrt{2}}{2},
    \label{eq:src_error}
\end{equation}
reaching a maximum of $9.90\,\mathrm{px}$ for $s{=}14$. This error is imprinted into the representation before any matching begins and cannot be removed by any subsequent matching strategy. On the target side, when the ground-truth location $\mathbf{y}$ falls between grid points, no candidate exists on the grid that represents its precise location. The set of unreachable keypoints on the target side is defined as:
\begin{equation}
    \mathcal{U}(\varepsilon) = \left\{\mathbf{y} \in \mathbb{R}^2 \;\middle|\; \min_{\mathbf{p} \in \mathcal{G}} \|\mathbf{y} - \mathbf{p}\| > \varepsilon \right\},
    \label{eq:unreachable}
\end{equation}
where $\varepsilon$ is the PCK threshold. Under the standard $448{\times}448$, patch-14 setting, 84.9\% of ground-truth keypoints fall into this set at PCK@0.01. This proportion is a structural ceiling already determined at the representation stage.

\paragraph{\textbf{Reformulation over a Continuous Feature Field.}}
To address both forms of quantization error at the representation level, we define a continuous feature field $F_{\mathbf{I}}: \mathbb{R}^2 \rightarrow \mathbb{R}^C$ for each image $\mathbf{I}$ that is queryable at arbitrary sub-pixel coordinates. Unlike grid features, which take a fixed value only at each grid point, $F_{\mathbf{I}}$ is trained to return a value that varies continuously within a grid region, conditioned on the grid feature — that is, querying any position between grid points yields a feature specific to that location. The concrete architecture of $F_{\mathbf{I}}$ is detailed in Section~\ref{sec:continuous_field}. For brevity, we write $F_s := F_{\mathbf{I}_s}$ and $F_t := F_{\mathbf{I}_t}$ hereafter.

Correspondence estimation is then reformulated over the continuous feature field as:
\begin{equation}
    \hat{\mathbf{y}} = \underset{\mathbf{q} \in \mathcal{G}'}{\arg\max}\; \mathrm{sim}\!\left(F_s(\mathbf{x}),\; F_t(\mathbf{q})\right),
    \label{eq:continuous_matching}
\end{equation}
where $\mathcal{G}'$ is a search grid $r$ times denser than the backbone grid $\mathcal{G}$, satisfying $|\mathcal{G}'| \gg |\mathcal{G}|$, and $\mathbf{q} \in \mathcal{G}'$ denotes a candidate coordinate on $\mathcal{G}'$. The distinction between Eq.~\eqref{eq:grid_matching} and Eq.~\eqref{eq:continuous_matching} is not merely one of resolution, but reflects a fundamental change in the nature of the representation.

On the source side, $F_s$ is queried directly at the exact keypoint coordinate $\mathbf{x}$, rather than at its nearest grid point $\hat{\mathbf{x}}$. Since the exact coordinate $\mathbf{x}$ is passed to $F_s$ without approximation, the representation error of Eq.~\eqref{eq:src_error} is theoretically eliminated: $e_\mathrm{src}(\mathbf{x}) = 0$. This follows from the fact that $F_s$ precisely computes the offset between $\mathbf{x}$ and $\hat{\mathbf{x}}$ and incorporates it internally, the mechanism of which is detailed in Section~\ref{sec:continuous_field}. That is, regardless of where within the grid region the source keypoint lies, $F_s(\mathbf{x})$ returns a feature conditioned on its exact location.

On the target side, the continuous field $F_t$ in principle supports infinite resolution, but directly solving for $\arg\max$ over the continuous domain is computationally intractable and would require iterative coordinate optimization. Instead, we decode $F_t$ onto $\mathcal{G}'$ via a single forward pass, constructing a finite candidate set while retaining the expressive power of the continuous representation and avoiding iterative inference. Since the density of $\mathcal{G}'$ can be set independently of the backbone architecture, the size of the unreachable set $\mathcal{U}(\varepsilon)$ in Eq.~\eqref{eq:unreachable} can be reduced without additional backbone computation.

Notably, interpolation cannot reduce the source-side quantization error to zero. We overcome this through explicit supervised learning of the continuous feature field: by training it to produce accurate correspondences at arbitrary coordinates, we directly model the per-location semantic variation that grid features alone cannot recover.

\subsection{Continuous Feature Field}
\label{sec:continuous_field}

We now present the concrete architecture that realizes the continuous feature field $F_{\mathbf{I}}: \mathbb{R}^2 \rightarrow \mathbb{R}^C$ introduced conceptually in Section~\ref{sec:formulation}. We propose a FiLM (Feature-wise Linear Modulation)-conditioned decoder~\cite{perez2018film}. FiLM is a modulation technique that generates channel-wise scale and shift parameters from a conditioning input to transform a feature representation. This decoder is shared across all images, and takes as input a grid feature and a relative offset from the grid center to decode the feature at an arbitrary query coordinate.

\paragraph{\textbf{FiLM-Conditioned Decoder.}}
The grid feature $\mathbf{Z}$ is first transformed into a latent code map via a $1 \times 1$ convolution, from which the latent code $\mathbf{z}^* = \mathbf{Z}[\hat{\mathbf{x}}]$ of each grid point is retrieved. The offset $\mathbf{\Delta} = \mathbf{x} - \hat{\mathbf{x}}$ between the query coordinate $\mathbf{x}$ and the grid point $\hat{\mathbf{x}}$ is projected into a $2C$-dimensional vector via a two-layer MLP $\varphi$:
\begin{equation}
    [\boldsymbol{\gamma};\, \boldsymbol{\beta}] = \varphi(\mathbf{\Delta}) \in \mathbb{R}^{2C},
    \label{eq:film_mlp}
\end{equation}
where $\boldsymbol{\gamma}, \boldsymbol{\beta} \in \mathbb{R}^C$ denote the channel-wise gain and bias, respectively. The latent code $\mathbf{z}^*$ is then modulated as:
\begin{equation}
    \mathbf{h} = \left(1 + \boldsymbol{\gamma}(\mathbf{\Delta})\right) \odot \mathbf{z}^* + \boldsymbol{\beta}(\mathbf{\Delta}),
    \label{eq:film_modulation}
\end{equation}
where $\odot$ denotes element-wise multiplication.

The grid feature $\mathbf{z}^*$ encodes the semantic content of an entire patch region into a single vector; however, even within the same patch, the semantic channels that should be emphasized vary depending on the precise query location. For instance, within a patch covering a bird's wing, the tip and the middle of the wing exhibit distinct semantic characteristics. By generating channel-wise gain $\boldsymbol{\gamma}(\mathbf{\Delta})$ and bias $\boldsymbol{\beta}(\mathbf{\Delta})$ from the offset $\mathbf{\Delta}$ and applying them to modulate $\mathbf{z}^*$, FiLM enables the decoder to produce different features from the same grid representation depending on the within-grid query location.

The modulated feature $\mathbf{h}$ is subsequently refined by a nonlinear rendering MLP $\rho$, and the per-grid-point decoder output $f_\theta$ is defined as:
\begin{equation}
    f_\theta(\mathbf{z}^*, \mathbf{\Delta}) := \rho(\mathbf{h}) = \mathrm{Linear}(\mathrm{ReLU}(\mathrm{Linear}(\mathrm{ReLU}(\mathbf{h})))),
    \label{eq:render_mlp}
\end{equation}
where $\rho$ is a width-preserving ($C \rightarrow C \rightarrow C$) two-layer network with a pre-activation design in which the leading ReLU acts directly on $\mathbf{h}$. This nonlinear rendering stage is central to the expressive power of the decoder, learning complex interactions between the offset $\mathbf{\Delta}$ and the grid feature $\mathbf{z}^*$ that the linear FiLM modulation alone cannot represent.

\begin{figure*}[t]
    \centering
    \includegraphics[width=0.85\textwidth]{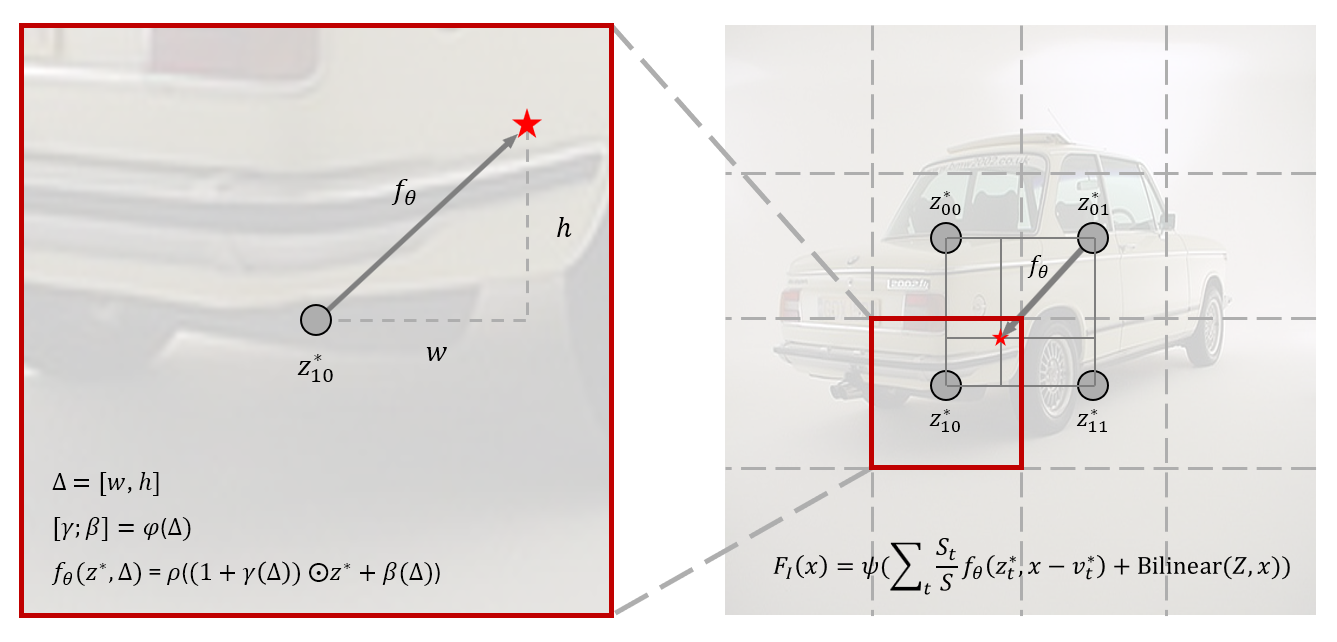}
    \caption{
        Overview of the FiLM-conditioned decoder and local ensemble.
        (Left) Zoomed-in view of the decoding process for grid point $\mathbf{z}^*_{10}$: the offset $\mathbf{\Delta} = [w, h]$ relative to the query coordinate is transformed by an MLP $\varphi$ into channel-wise modulation parameters $[\boldsymbol{\gamma}; \boldsymbol{\beta}]$ (Eq.~\eqref{eq:film_mlp}), which modulate the grid feature (Eq.~\eqref{eq:film_modulation}) and are subsequently refined by a nonlinear rendering MLP $\rho$ to produce the per-grid-point feature $f_\theta(\mathbf{z}^*, \mathbf{\Delta})$ (Eq.~\eqref{eq:render_mlp}).
        (Right) The four per-grid-point features surrounding the query coordinate (red star) are combined via area-based weighting, added to a bilinear residual, and projected through a linear layer $\psi$ to produce the final feature $F_{\mathbf{I}}(\mathbf{x})$ (Eq.~\eqref{eq:local_ensemble}).
    }
    \label{fig:method}
\end{figure*}

\paragraph{\textbf{Local Ensemble and Final Projection.}}
Evaluating $f_\theta$ solely at the single nearest grid point introduces discontinuities at the boundaries between adjacent grid regions — as the query coordinate crosses a grid boundary, $\mathbf{z}^*$ abruptly switches to the feature of a neighboring grid point. Inspired by works that ensure continuity by aggregating contributions from neighboring grid points~\cite{chen2021learning}, we adopt a local ensemble. For the four grid points $\{(\mathbf{z}^*_t, \mathbf{v}^*_t)\}_{t \in \{00, 01, 10, 11\}}$ surrounding the query coordinate $\mathbf{x}$, $f_\theta(\mathbf{z}^*_t, \mathbf{x} - \mathbf{v}^*_t)$ is computed independently at each, and the results are combined using the area $S_t$ subtended between $\mathbf{x}$ and the diagonally opposite grid point of each neighbor as weights. For training stability, a bilinear interpolation of the original grid feature $\mathbf{Z}$ at the query coordinate $\mathbf{x}$ is added to the ensembled feature as a residual, and the result is finally projected to the output feature dimension via a linear layer $\psi$, completing $F_{\mathbf{I}}(\mathbf{x})$:
\begin{equation}
    F_{\mathbf{I}}(\mathbf{x}) = \psi\!\left( \sum_{t} \frac{S_t}{S} \cdot f_\theta\!\left(\mathbf{z}^*_t,\; \mathbf{x} - \mathbf{v}^*_t\right) + \mathrm{Bilinear}(\mathbf{Z}, \mathbf{x}) \right),
    \label{eq:local_ensemble}
\end{equation}
where $S = \sum_t S_t$ and $\mathrm{Bilinear}(\mathbf{Z}, \mathbf{x})$ denotes the bilinear interpolation of the original grid feature $\mathbf{Z}$ at $\mathbf{x}$. As $\mathbf{x}$ approaches grid point $\mathbf{v}^*_t$, the corresponding weight $S_t$ increases, ensuring that the feature field transitions smoothly across grid boundaries. This decoder is applied symmetrically on both sides of correspondence estimation: on the source side, it is queried at the exact keypoint coordinate $\mathbf{x}$ to obtain $F_s(\mathbf{x})$; on the target side, it is evaluated at every candidate $\mathbf{q} \in \mathcal{G}'$ to obtain $F_t(\mathbf{q})$. Since the offset $\mathbf{\Delta} = \mathbf{x} - \hat{\mathbf{x}}$ used in Eq.~\eqref{eq:film_modulation} is computed exactly, without approximation, and incorporated into the modulation, $F_s(\mathbf{x})$ realizes $e_\mathrm{src}(\mathbf{x}) = 0$ as claimed in Section~\ref{sec:formulation}.

\subsection{Training and Inference}
\label{sec:training}

\paragraph{\textbf{Training.}}
The objective of training is to learn a continuous feature field $F_{\mathbf{I}}$ that faithfully represents the semantic feature at any queried coordinate, such that it can produce accurate correspondences at arbitrary sub-pixel locations. To this end, the backbone encoder is adapted from its pretrained weights via a lightweight parameter-efficient method~\cite{hu2022lora}, and is trained jointly with the FiLM-conditioned decoder.

During training, a source-target image pair $(\mathbf{I}_s, \mathbf{I}_t)$ is provided along with ground-truth keypoint coordinate pairs $(\mathbf{x}, \mathbf{y})$. The resulting features $F_s(\mathbf{x})$ and $\{F_t(\mathbf{q})\}_{\mathbf{q} \in \mathcal{G}'}$ yield a predicted correlation distribution $P$ over $\mathcal{G}'$ via softmax over $\mathrm{sim}(F_s(\mathbf{x}), F_t(\mathbf{q}))$, where $(i,j)$ indexes a candidate location on $\mathcal{G}'$.

Following MARCO~\cite{cuttano2026marco}, we supervise the predicted correlation distribution $P$ with a Gaussian soft target $g_\sigma$ centered at the ground-truth location, annealing $\sigma$ from $\sigma_{\max}$ to $\sigma_{\min}$ over training to realize a coarse-to-fine curriculum:
\begin{equation}
    \mathcal{L}_{CE} = -\sum_{(i,j)} g_\sigma(i,j) \log P(i,j).
    \label{eq:loss}
\end{equation}

\paragraph{\textbf{Inference.}}
At inference time, window soft-argmax~\cite{zhang2024telling} is applied within a local window $\mathcal{W}(\hat{\mathbf{y}}_0)$ centered at the initial estimate $\hat{\mathbf{y}}_0$ obtained from Eq.~\eqref{eq:continuous_matching}, producing the final sub-pixel correspondence:
\begin{equation}
    \hat{\mathbf{y}} = \sum_{\mathbf{q} \in \mathcal{W}(\hat{\mathbf{y}}_0)} \mathbf{q} \cdot \mathrm{softmax}\!\left(\mathrm{sim}(F_s(\mathbf{x}),\, F_t(\mathbf{q})) / \tau_{\mathrm{inf}}\right).
    \label{eq:soft_argmax}
\end{equation}
This final step further reduces the residual quantization error introduced by the discrete grid $\mathcal{G}'$, yielding a continuous correspondence estimate.

\section{Experiments}
\subsection{Implementation Details}
\label{sec:implementation}

We use pretrained DINOv2-B/14~\cite{oquab2023dinov2} as the backbone, kept frozen and adapted with LoRA~\cite{hu2022lora}, with input images resized so that the feature map is $64 \times 64$. Training uses batch size 4 and Adam~\cite{kingma2014adam} with an initial learning rate of $6 \times 10^{-4}$; the Gaussian soft-target width $\sigma$ is annealed from 3 to 1. On the target side we use a densification factor of $r = 4$, and at inference window soft-argmax is applied within a window of size 45. Full training details, including the LoRA configuration and training cost, are given in the supplementary material.

\subsection{Datasets and Evaluation Metric}
\label{sec:datasets}

\paragraph{\textbf{Datasets.}}
\noindent{\textbf{SPair-71k}}~\cite{min2019spair} is a large-scale semantic correspondence benchmark comprising 70,958 image pairs across 18 object categories. It evaluates correspondence between diverse object instances within the same category, providing a challenging setting with substantial variations in viewpoint, scale, and appearance. Its large scale and categorical diversity enable a thorough assessment of the generalization ability of correspondence methods.

\noindent{\textbf{AP-10K}}~\cite{yu2021ap} is a large-scale animal pose estimation benchmark comprising 10,015 images spanning 23 animal families and 54 species. For semantic correspondence evaluation, it is organized into three splits of increasing difficulty — intra-species, cross-species, and cross-family — each introducing progressively larger appearance and structural variation. These splits enable a rigorous evaluation of generalization ability in a domain distinct from SPair-71k.

\paragraph{\textbf{Evaluation Metric.}}
We evaluate using Percentage of Correct Keypoints (PCK), the standard metric for semantic correspondence. A predicted keypoint is considered correct if it falls within a radius of $\alpha \cdot \max(h, w)$ from the ground-truth keypoint, where $h$ and $w$ denote the height and width of the target object bounding box on both SPair-71k and AP-10K ($\alpha_\text{bbox}$). We report results at three thresholds $\alpha \in \{0.01, 0.05, 0.1\}$, with particular emphasis on performance at the fine-grained thresholds PCK@0.01 and PCK@0.05.

\begin{table*}[t]
\centering
\scriptsize
\setlength{\tabcolsep}{2.2pt}
\caption{Quantitative comparison on standard semantic correspondence benchmarks. We report per-image PCK (\%, $\uparrow$) at multiple thresholds on SPair-71k and three splits of AP-10K (intra-species, cross-species, cross-family). The best and second-best results per column are \textbf{bolded} and \underline{underlined}, respectively. $\S$ uses depth maps at training; $\dagger$ uses object masks at training; $\ddagger$ uses object masks at inference.}
\label{tab:main}

\resizebox{\textwidth}{!}{%
\begin{tabular}{l@{\hspace{10pt}}c|ccc|ccc|ccc|ccc}
\toprule
\multicolumn{2}{c}{} & \multicolumn{3}{c}{\textbf{SPair-71k}} & \multicolumn{3}{c}{\textbf{AP-10K (I.S.)}} & \multicolumn{3}{c}{\textbf{AP-10K (C.S.)}} & \multicolumn{3}{c}{\textbf{AP-10K (C.F.)}} \\
\cmidrule(lr){3-5} \cmidrule(lr){6-8} \cmidrule(lr){9-11} \cmidrule(lr){12-14}
Method & Backbone & 0.01 & 0.05 & 0.10 & 0.01 & 0.05 & 0.10 & 0.01 & 0.05 & 0.10 & 0.01 & 0.05 & 0.10 \\
\midrule
\multicolumn{14}{l}{\textit{Unsupervised/Weakly Supervised}} \\

DINOv2+NN~\cite{zhang2023tale} & ViT-B & 6.3 & 38.4 & 53.9 & 6.4 & 41.0 & 60.9 & 5.3 & 37.0 & 57.3 & 4.4 & 29.4 & 47.4 \\
DIFT~\cite{tang2023emergent} & SD & 7.2 & 39.7 & 52.9 & 6.2 & 34.8 & 50.3 & 5.1 & 30.8 & 46.0 & 3.7 & 22.4 & 35.0 \\
SD+DINO~\cite{zhang2023tale} & SD+ViT-B & 7.9 & 44.7 & 59.9 & 7.6 & 43.5 & 62.9 & 6.4 & 39.7 & 59.3 & 5.2 & 30.8 & 48.3 \\
DIY-SC~\cite{dunkel2025yourself} & SD+ViT-B & 10.1 & 53.8 & 71.6 & - & - & 70.6 & - & - & 69.8 & - & - & 57.8 \\
\midrule
\multicolumn{14}{l}{\textit{Supervised methods}} \\

NeMF~\cite{hong2022neural} & ResNet101 & 3.2 & 34.2 & 53.6 & - & - & - & - & - & - & - & - & - \\
DHF~\cite{luo2023diffusion} & SD & 8.7 & 50.2 & 64.9 & 8.0 & 45.8 & 62.7 & 6.8 & 42.4 & 60.0 & 5.0 & 32.7 & 47.8 \\
SD+DINO (S)~\cite{zhang2023tale} & SD+ViT-B & 9.6 & 57.7 & 74.6 & 9.9 & 57.0 & 77.0 & 8.8 & 53.9 & 74.0 & 6.9 & 46.2 & 65.8 \\
GECO$^\dagger$~\cite{hartwig2025geco} & ViT-B & 14.2 & 59.6 & 73.6 & 19.2 & 67.1 & 82.5 & 17.4 & 64.9 & 81.2 & 14.5 & 60.4 & 76.6 \\
Jamais Vu$^\dagger$$^\S$~\cite{mariotti2025jamais} & SD+ViT-B & 20.5 & 71.9 & 82.5 & - & - & - & - & - & - & - & - & - \\
Geo-SC$^\ddagger$~\cite{zhang2024telling} & SD+ViT-B & 21.7 & 72.8 & 83.2 & 23.2 & 73.2 & 87.7 & 21.7 & 70.3 & 85.9 & 18.3 & 63.2 & 78.5 \\
MARCO$^\dagger$~\cite{cuttano2026marco} & ViT-L & \underline{27.0} & \underline{77.6} & \textbf{87.2} & \underline{32.6} & \underline{77.4} & \underline{89.1} & \underline{32.2} & \underline{76.6} & \underline{88.3} & \underline{28.5} & \underline{71.1} & \textbf{83.4} \\

\textbf{ImCorr} (Ours)                        & ViT-B  & \textbf{33.2} & \textbf{78.9} & \underline{86.2} & \textbf{36.2} & \textbf{78.6} & \textbf{90.0} & \textbf{35.7} & \textbf{77.7} & \textbf{89.4} & \textbf{32.4} & \textbf{73.6} & \underline{82.9} \\
\bottomrule
\end{tabular}%
}
\end{table*}

\subsection{Comparison with State-of-the-Art}
\label{sec:sota}
 
Table~\ref{tab:main} presents the quantitative comparison on SPair-71k~\cite{min2019spair} and AP-10K (intra-species, cross-species, and cross-family)~\cite{yu2021ap}. ImCorr achieves a new state of the art at fine-grained thresholds (PCK@0.01, PCK@0.05) across all four benchmarks. At PCK@0.01, ImCorr improves over the previous best-performing method, MARCO~\cite{cuttano2026marco}, by $+6.2$\%p on SPair-71k, $+3.6$\%p on AP-10K intra-species, $+3.5$\%p on cross-species, and $+3.9$\%p on cross-family, while also ranking first on every benchmark at PCK@0.05. The margin over the runner-up methods excluding MARCO, namely Geo-SC~\cite{zhang2024telling} and Jamais Vu~\cite{mariotti2025jamais}, is considerably larger: on SPair-71k at PCK@0.01, ImCorr outperforms Geo-SC by $+11.5$\%p and Jamais Vu by $+12.7$\%p.
 
At the standard threshold (PCK@0.10), ImCorr also maintains competitive performance with existing state-of-the-art methods across SPair-71k, AP-10K intra-species, cross-species, and cross-family.
 
Notably, these gains are achieved with fewer resources. ImCorr employs DINOv2-B/14~\cite{oquab2023dinov2} as its backbone, whereas MARCO relies on the larger DINOv2-L/14. Furthermore, while comparison methods including Geo-SC, MARCO, and Jamais Vu leverage additional supervisory signals such as object masks or depth maps during training or inference, ImCorr relies solely on keypoint annotations, yet surpasses all of them at fine-grained thresholds.

\begin{figure*}[t]
    \centering
    \includegraphics[width=\textwidth]{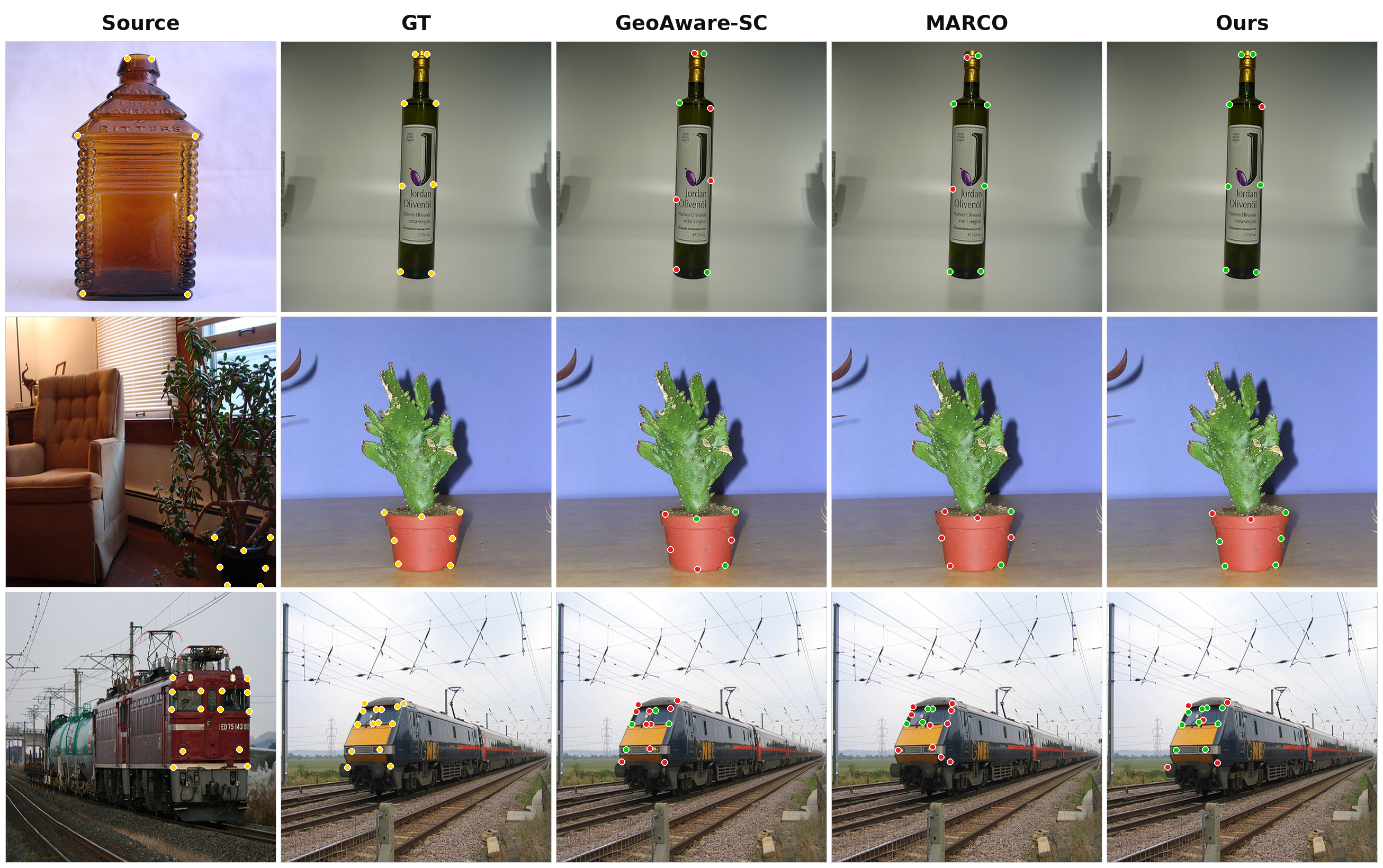}
    \caption{
        Qualitative comparison on SPair-71k (PCK@0.01, per-image). Green and red dots indicate correct and incorrect correspondences, respectively. While GeoAware-SC and MARCO capture the approximate location of most keypoints, they tend to deviate slightly from the ground truth under the fine-grained threshold. In contrast, ImCorr matches the same keypoints with substantially higher precision.
    }
    \label{fig:vis2}
\end{figure*}

Figure~\ref{fig:vis2} shows qualitative comparisons on SPair-71k under the PCK@0.01 threshold. GeoAware-SC~\cite{zhang2024telling} and MARCO~\cite{cuttano2026marco} capture the approximate location of most keypoints, but under the fine-grained threshold, their predictions deviate slightly from the ground truth, resulting in a substantial number of incorrect matches shown in red. In contrast, ImCorr achieves a higher proportion of correct matches, shown in green, on the same keypoints, and accurately localizes keypoints even in densely clustered regions such as bottle necks and train fronts.

\subsection{Ablation Study}

\paragraph{\textbf{Source-Side Query and Target-Side Decoding.}}
Table~\ref{tab:ablation_src_tgt} analyzes the individual contributions of source-side continuous query and target-side dense decoding. The baseline corresponds to a purely grid-based matching method that directly uses backbone grid features without the FiLM decoder.
 
The full model achieves gains of $+14.1$, $+6.0$, and $+2.9$ percentage points over the baseline at PCK@0.01, PCK@0.05, and PCK@0.10, respectively. The monotonically increasing improvement as the threshold decreases suggests that the two components effectively address the precision loss induced by grid quantization. The nature of this improvement, however, differs markedly between the two components.
 
Source-side continuous query yields a gain of $+6.5$\%p at PCK@0.01 while also achieving a meaningful improvement of $+3.4$\%p at PCK@0.10. This indicates that querying features at the exact keypoint coordinate improves not only fine-grained precision but also the overall quality of the matching representation. In contrast, target-side dense decoding achieves the largest gain at PCK@0.01 ($+11.0$\%p) while contributing only marginally at PCK@0.10 ($+0.8$\%p). This confirms that expanding the resolution of the search space substantially reduces the proportion of unreachable candidates at fine-grained thresholds, while the existing grid resolution is already sufficient at standard thresholds.
 
The two components address quantization error through complementary mechanisms — the source side improves the precision of the feature representation, while the target side expands the resolution of the search space. When combined, the full model achieves PCK@0.01 of 33.2\%, exhibiting complementary synergy that exceeds the sum of the individual gains ($+6.5$\%p and $+11.0$\%p), and confirming that the two components are mutually reinforcing rather than redundant.

\begin{table}[ht]
\centering
\small
\setlength{\tabcolsep}{6pt}
\caption{Ablation on source-side query and target-side decoding. We report per-image PCK (\%, $\uparrow$) on SPair-71k. \checkmark denotes the component is enabled.}
\label{tab:ablation_src_tgt}
\begin{tabular}{l@{\hspace{8pt}}cc@{\hspace{10pt}}ccc}
\toprule
Setting & Src & Tgt & PCK@0.01 & PCK@0.05 & PCK@0.10 \\
\midrule
Baseline              &            &              & 19.1 & 72.9 & 83.3 \\
+ Src Query           & \checkmark &              & 25.6  & 77.2 & \textbf{86.7} \\
+ Tgt Decoding        &            & \checkmark   & 30.1 & 76.1 & 84.1 \\
Full model (Ours) & \checkmark & \checkmark & \textbf{33.2} & \textbf{78.9} & 86.2 \\
\quad{\scriptsize Gain over baseline} & & &
{\scriptsize\textcolor{green!50!black}{+14.1}} &
{\scriptsize\textcolor{green!50!black}{+6.0}} &
{\scriptsize\textcolor{green!50!black}{+2.9}} \\
\bottomrule
\end{tabular}
\end{table}

\paragraph{\textbf{Densification Factor $r$.}}
Table~\ref{tab:ablation_r} analyzes the effect of the densification factor $r$ of the dense search grid $\mathcal{G}'$. As $r$ increases from 2 to 4, consistent improvements are observed at fine-grained thresholds: PCK@0.01 increases from 30.3 to 33.2, and PCK@0.05 from 76.2 to 78.9. This is attributable to the fact that a denser grid increases the likelihood that ground-truth locations previously unreachable within the candidate set become accessible. Beyond $r=4$, however, PCK@0.01 decreases monotonically to 32.9 and 31.7 at $r=6$ and $r=8$, respectively. We interpret this as a consequence of excessive densification, where an abundance of candidates with similar features degrades the discriminability of the matching.

At PCK@0.10, performance decreases monotonically from 86.9 to 85.1 as $r$ increases, a pattern that is consistent with the findings of Table~\ref{tab:ablation_src_tgt}. Just as target-side dense decoding contributed only marginally to PCK@0.10 ($+0.8$\%p) in the component ablation, increasing $r$ similarly fails to improve standard-threshold performance and instead induces a slight degradation. This consistently suggests that the backbone grid resolution is already sufficient at standard thresholds, and that the primary effect of densification lies in expanding the search space at fine-grained thresholds.

\begin{table}[ht]
\centering
\small
\setlength{\tabcolsep}{9pt}
\caption{Ablation on the densification  factor $r$ of the dense search grid $\mathcal{G}'$. We report per-image PCK (\%, $\uparrow$) on SPair-71k.}
\label{tab:ablation_r}
\begin{tabular}{c@{\hspace{14pt}}|@{\hspace{14pt}}ccc}
\toprule
$r$ & PCK@0.01 & PCK@0.05 & PCK@0.10 \\
\midrule
2   & 30.3        & 76.2        & \textbf{86.9}        \\
4   & \textbf{33.2} & \textbf{78.9} & 86.2 \\
6   & 32.9        & 78.0        & 85.6        \\
8   & 31.7        & 77.1        & 85.1        \\
\bottomrule
\end{tabular}
\end{table}

\paragraph{\textbf{Inference Resolution.}}
To ensure a fair comparison, we additionally evaluate ImCorr at an input resolution of 770, matching the inference resolution of MARCO~\cite{cuttano2026marco}, which holds the previous state of the art at fine-grained thresholds. Despite using a backbone that is 3.5$\times$ smaller than that of MARCO (DINOv2-B, 86M vs.\ DINOv2-L, 303M parameters)~\cite{oquab2023dinov2}, ImCorr achieves 32.1\% at PCK@0.01 on SPair-71k under this setting, outperforming MARCO (27.0\%) by 5.1\%p. This confirms that the performance gains stem from the structural design of the continuous feature field, rather than from differences in input resolution or backbone capacity.

\paragraph{\textbf{Interpolation Baseline.}}
To directly validate our claim that a continuous field cannot be substituted by interpolation, we additionally evaluate a variant of the baseline in Table~\ref{tab:ablation_src_tgt} (a purely grid-based method that directly uses backbone grid features without the FiLM decoder) equipped with bilinear interpolation on both the source and target sides. This variant achieves 20.9\% at PCK@0.01 on SPair-71k, a modest improvement over the baseline (19.1\%) but still substantially below our full model (33.2\%). This confirms that interpolation merely blends existing grid values without generating new position-specific semantic information, whereas the learned FiLM decoder directly learns to produce it.

\section{Limitation and Future Work}
\label{sec:limitation}

While ImCorr enables feature queries at arbitrary sub-pixel coordinates, several limitations remain. The target-side quantization error is only substantially reduced through densification rather than eliminated, since $\mathcal{G}'$ remains a finite discrete grid.

More fundamentally, our decoder conditions on the relative offset of a query point through a single FiLM modulation, treating every location within a cell as a flat, axis-aligned coordinate. It therefore cannot represent where a query lies within a patch hierarchically, from coarse to fine, and its axis-aligned parameterization is misaligned with the orientation-agnostic, radius-based criterion that PCK employs. As a result, precise localization within a single patch remains an open problem, and accuracy at the strictest threshold is still far below that at the standard threshold. Encoding within-cell position in a more structured, geometry-aware manner is a promising direction for future work.

\section{Conclusion}

There are no grid lines in nature, yet the models that estimate semantic correspondence have consistently searched for answers only on a grid. This paper has shown that this gap is the fundamental bottleneck of precise correspondence, and has proposed ImCorr, which directly resolves this at the representation level by defining correspondence over a continuous feature field rather than a discrete grid. Experiments on SPair-71k and AP-10K demonstrate that ImCorr achieves consistent state-of-the-art performance at fine-grained thresholds (PCK@0.01, PCK@0.05). These results suggest that the bottleneck of precise correspondence has long been obscured behind the loose thresholds of standard benchmarks, and that representational continuity is its effective solution.

%
%
\bibliographystyle{splncs04}
\bibliography{main}

\appendix

\section{Implementation Details}

\paragraph{Backbone and adaptation.}
We use pretrained DINOv2-B/14~\cite{oquab2023dinov2} as the backbone encoder. Input images are resized such that the resulting feature map has a spatial resolution of $64 \times 64$ (an input resolution of $896 \times 896$). The backbone weights are kept frozen and adapted with LoRA~\cite{hu2022lora} (rank 16, $\alpha = 1$, i.e., scaling $\alpha / \text{rank} = 1/16$), applied to the second MLP projection (\texttt{fc2}) of the last six transformer blocks (blocks 6--11, 0-indexed).

\paragraph{Decoder.}
The offset MLP $\varphi$ has a hidden dimension of 64. The rendering MLP $\rho$ is width-preserving ($C \rightarrow C \rightarrow C$) with a pre-activation design, as described in Sec.~3.2.

\paragraph{Training.}
Training uses a batch size of 4 and the Adam optimizer~\cite{kingma2014adam} with an initial learning rate of $6 \times 10^{-4}$ and a StepLR scheduler. The standard deviation $\sigma$ of the Gaussian soft target is annealed from $\sigma_{\max} = 3$ to $\sigma_{\min} = 1$ over the course of training. Training runs for 5 epochs on a single NVIDIA A100 SXM, taking approximately 30 hours.

\paragraph{Inference.}
On the target side, a densification factor of $r = 4$ is applied to the dense search grid $\mathcal{G}'$, yielding a $256 \times 256$ candidate lattice. Window soft-argmax is applied within a local window of size 45 to produce the final sub-pixel correspondence.

\section{Additional Qualitative Results}
We present additional qualitative results on SPair-71k at PCK@0.01 across a broader range of object categories (green: correct, red: incorrect). ImCorr consistently produces higher-precision correspondences than existing methods, even in densely clustered keypoint regions.

\begin{figure*}[ht]
    \centering
    \includegraphics[width=0.9\textwidth]{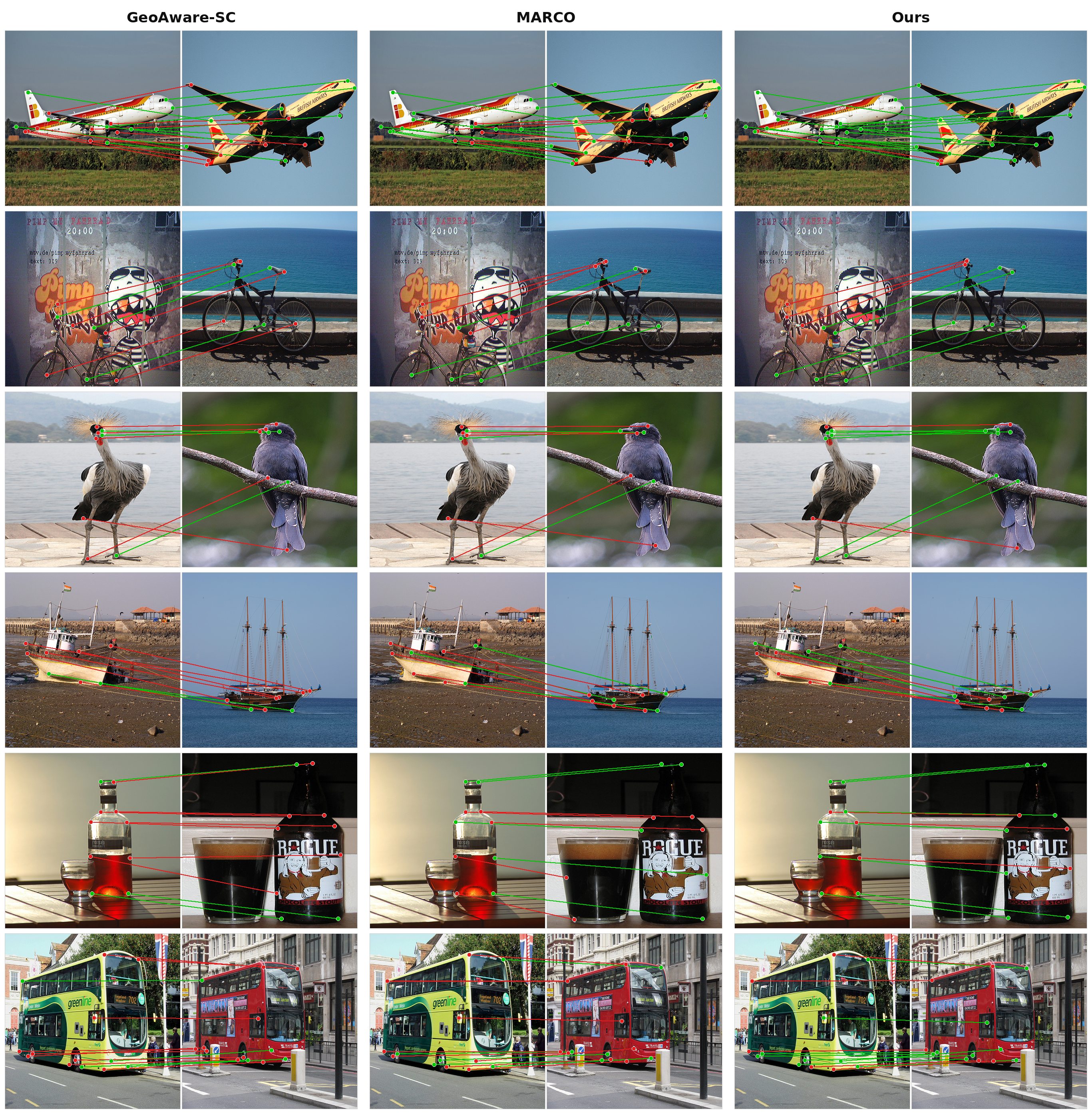}
    \caption{
        Additional results on SPair-71k across object categories.
    }
    \label{fig:appendix_qualitative_1}
\end{figure*}

\begin{figure*}[p]
    \vspace*{\fill}
    \centering
    \includegraphics[width=0.9\textwidth]{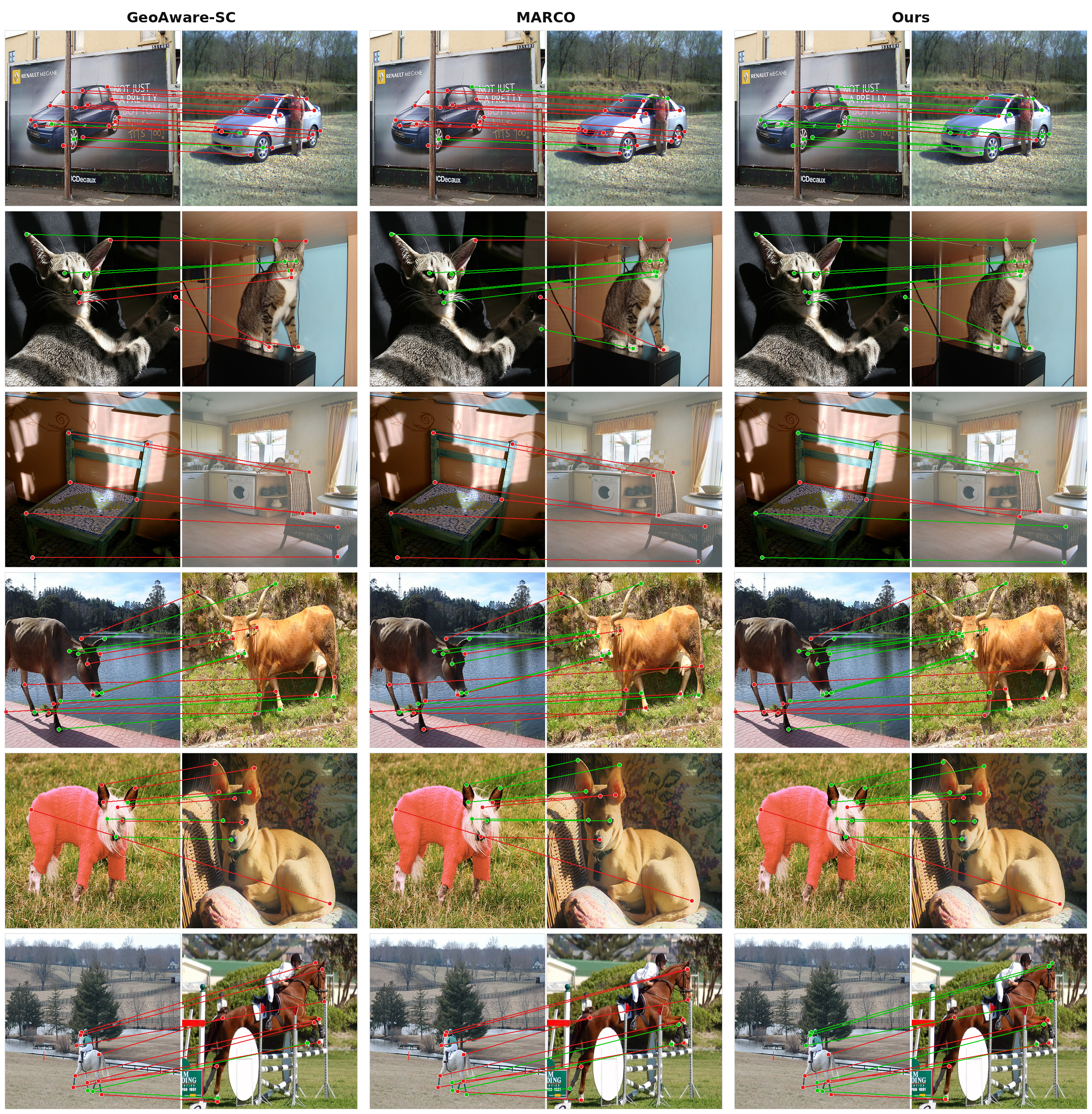}
    \caption{
        Additional results on SPair-71k across object categories.
    }
    \label{fig:appendix_qualitative_2}
    \vspace*{\fill}
\end{figure*}

\begin{figure*}[ht]
    \centering
    \includegraphics[width=0.9\textwidth]{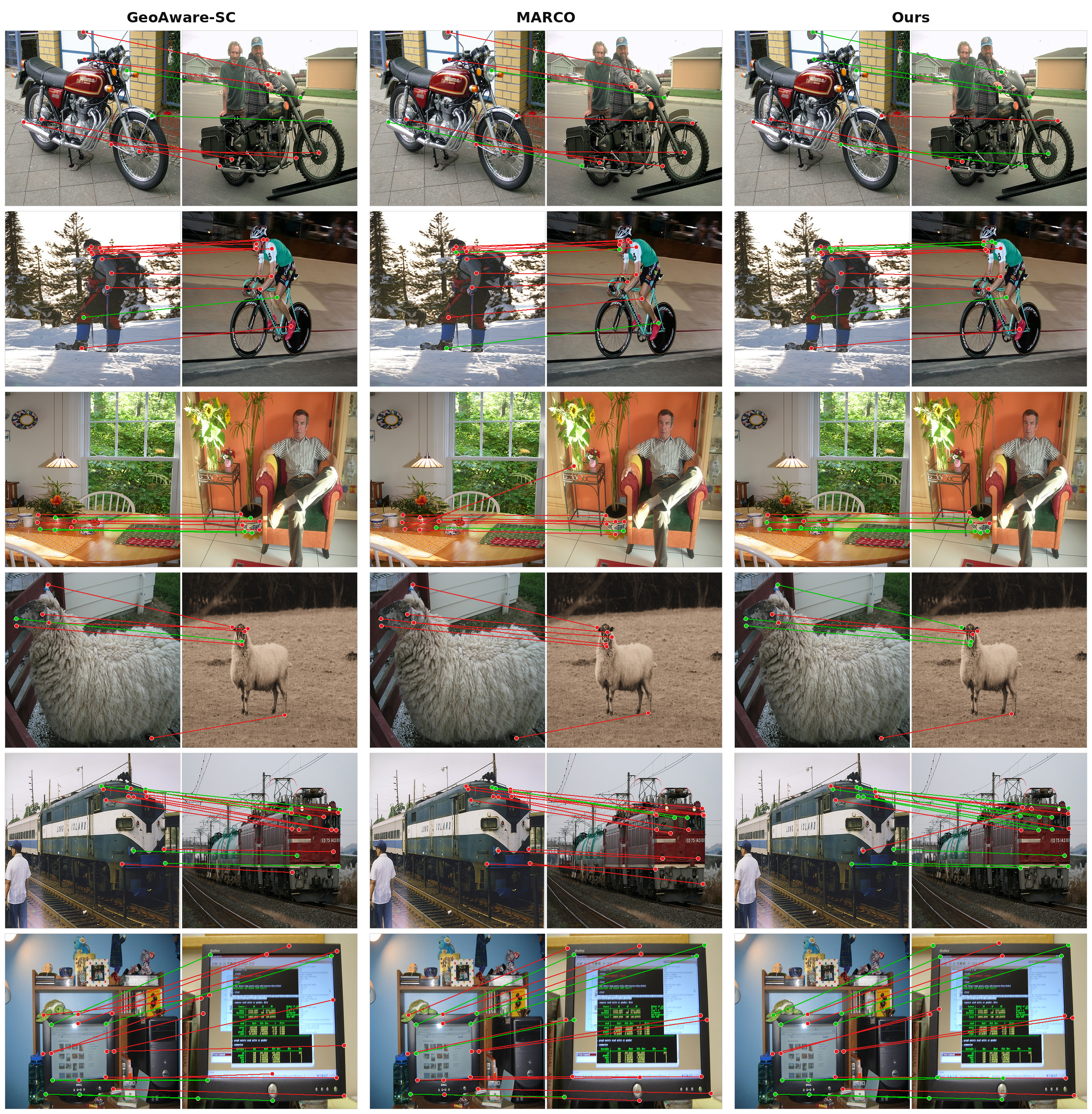}
    \caption{
        Additional results on SPair-71k across object categories.
    }
    \label{fig:appendix_qualitative_3}
\end{figure*}

\end{document}